\documentclass{llncs}
\usepackage{graphicx}
\usepackage{color}
\usepackage{bm}
\usepackage{amsmath}
\usepackage{amssymb}
\usepackage{setspace}
\usepackage{array}
\usepackage{tabularx}
\usepackage{multirow}
\usepackage{afterpage}
\usepackage{hhline}
\usepackage{url}
\usepackage{hyperref} 
\usepackage{footmisc} 
\usepackage{xspace}
\usepackage[numbers,sort&compress]{natbib}
\usepackage{import}
\usepackage{caption}
\usepackage{subcaption}
\usepackage{bibentry}
\nobibliography*

\newcommand{\comment}[1]{}

\providecommand{\e}[1]{\ensuremath{\times 10^{#1}}}

\title{Exploring the power of GPU's for training Polyglot language models}
\author{Vivek Kulkarni, Rami Al-Rfou', Bryan Perozzi, Steven Skiena}
\institute{Department of Computer Science\\
Stony Brook University\\
\{\email{bperozzi,ralrfou,vvkulkarni,skiena}\}\email{@cs.stonybrook.edu}}

\begin{document}
\maketitle
\begin{abstract}
One of the major research trends currently is the evolution of heterogeneous parallel computing. GP-GPU computing is being widely used and several applications have been designed to exploit the massive parallelism that GP-GPU's have to offer. While GPU's have always been widely used in areas of computer vision for image processing, little has been done to investigate whether the massive parallelism provided by GP-GPU's can be utilized effectively for Natural Language Processing(NLP) tasks. In this work, we investigate and explore the power of GP-GPU's in the task of learning language models. More specifically, we investigate the performance of training Polyglot language models\cite{polyglot} using deep belief neural networks. We evaluate the performance of training the model on the GPU and present optimizations that boost the performance on the GPU.One of the key optimizations, we propose increases the performance of a function involved in calculating and updating the gradient by approximately 50 times on the GPU for sufficiently large batch sizes. We show that with the above optimizations, the GP-GPU's performance on the task increases by factor of approximately 3-4. The optimizations we made are generic Theano\cite{bergstra+al:2010-scipy} optimizations and hence potentially boost the performance of other models which rely on these operations.We also show that these optimizations result in the GPU's performance at this task being now comparable to that on the CPU. We conclude by presenting a thorough evaluation of the applicability of GP-GPU's for this task and highlight the factors limiting the performance of training a Polyglot model on the GPU.
\end{abstract}
\section{Introduction}
\label{sec:Introduction}
GPU computing has been of significant interest to the research community. This has led to the evolution of heterogeneous parallel computing with the realization that clock speeds on CPU's have reached their
limit. Heterogeneous Parallel Computing offers a solution to this problem by offloading computation that can be massively parallelized on the GPU's and running serial computation on the CPUs. Thus GP-GPU applications are widely used to solve several computational problems. For example, BarraCUDA is a software that uses GPU's to speed up the alignment of short sequence reads to a particular location on some
reference genome\cite{Doe:2009:Misc}. GPU's have always been extensively used in the field of computer vision for image processing. To illustrate, OpenVidea is a system that implements several computer vision algorithms on the GPU\cite{Fung:2005:OPG:1101149.1101334}.It is to be noted, that most images are represented as dense matrices, and image processing algorithms render themselves to be effectively parallelized as they tend to operate on stencils(a fixed template of pixels). Since GPU's are particularly suited for vector computations, image processing applications typically are designed to utilize the massive parallelism offered by the GPU's.

In contrast in the field of Natural Language processing, the models are typically sparse. This implies that language models have a sparse representation. Since the model is sparse, the computation does not
admit to being effectively accomplished on the GPU. However recent advances in machine learning and natural language processing have resulted in the development of new language models that learn representations\cite{Hinton:2006:FLA:1161603.1161605}\cite{Hinton28072006}. 
These representations which are also termed \emph{embeddings }are a dense representation. One popular technique of learning word representations is to use deep belief networks. Deep Belief networks are neural networks which tend to have a large number of hidden layers. Training deep belief networks requires massive amounts of training examples and thus requires massive computation. Deep belief networks used for solving computer vision tasks are most efficiently trained on the GPU\cite{NIPS2012_4824}. In this work, we investigate the performance of training such a language model(learnt by deep belief network) on the GPU. The language model we used for learning word representaions is very similar to the language model described in the \emph{SENNA} system\cite{collobert-2011}. We call our language model Polyglot. Polyglot allows one to learn word representations from massive amounts of unannotated text. We have used Polyglot to learn word embeddings for more than 100 different languages\footnote{Available: \url{http://bit.ly/embeddings}}. A detailed description of the Polyglot project is available in \cite{polyglot}.At a high level, we measure the performance of training a Polyglot model in terms of the number of training examples processed per second. We evaluate the performance in terms of the speed-up achieved on the GPU and compare it against the CPU. We next briefly describe our experimental setup and describe our research methodology.

\section{Experimental Setup}
\label{sec:Setup}
We run all our experiments on the GPU on GEForce GT $570$ . This GPU has $480$ cores, with a processor clock speed of $1464$MHz and a memory clock speed of $1900$MHz. GPU's in the GEForce family are
particularly suited for research applications. To train the language  model, we use a well-known library called \emph{Theano}\cite{bergstra+al:2010-scipy}. Theano is a symbolic computation library that is particularly suited for machine learning applications. It allows for developers to succinctly specify how the parameters of the model are to be calculated. Using Theano to learn a model, allows the developer to implement a machine
learning algorithm easily as the complexity involved in explicitly calculating the gradients of the loss function is abstracted out by Theano. Theano has functionality built in to calculate gradients and
perform mathematical computation easily. Theano also transparently offloads computation on to the GPU if required. Since training a model, significantly involves computation required to learn a set of parameters
that minimize a loss function, it is imperative that the functions implemented in Theano are efficient. This involves performing both macro level optimizations in optimizing the computational graph and
micro level optimizations that target specific functions. Having described our experimental setup, we now describe at a high level, our research methodology in identifying bottle necks and optimizing these bottle
necks.

\section{Research methodology}
\label{sec:Research}
We follow the below standard procedure to analyze and evaluate the
performance bottlenecks: 
\begin{enumerate}
\item Note the number of training examples processed per second on the CPU
and GPU to establish a baseline.
\item Use a profiler to identify top hot spots. 
\item Focus on the top hot spots and investigate how they can be optimized

\begin{enumerate}
\item Identify any computational graph optimizations possible to reduce
computation 
\item Optimize function calls by investigating how parallelism can be boosted. 
\item Micro-optimize function calls 
\item Unit Test the changes to ensure correctness 
\end{enumerate}
\item Repeat Steps 1 to 3 if required. 
\end{enumerate}

\section{Results}
\label{sec:Results}
\subsection{Baseline}
\label{sec:Baseline}
In this section, we present the baseline numbers for the GPU and the
CPU. On the CPU, the mean training rate was $5512.6$ examples/second($\sigma=30.315)$.
On the GPU, the mean training rate was $1265.8$ examples/second($\sigma=20.604$).

The goal is to investigate the performance on the GPU and evaluate
the bottle-necks involved. On scanning the logs, we note that the
percentage of total time approximately spent on Theano processing
was $96\%$.

This implies that it would be fruitful to analyze and focus on optimizations
in Theano to help boost performance. In the next section, we analyze
the performance of Theano using the built in Theano Profiler.

\subsection{Profiling Theano}
\label{sec:Profiling}
We profiled Theano to get some insight into what the performance hot
spots are, so that our efforts could be channeled into optimizing
those hot spots. We show the time taken per call and the fraction
of time spent for these functions in the Table 1 .We note that there
is 1 major hot spot, namely \emph{GpuAdvancedIncSubTensor1}.

\begin{table}
\label{Table1}\caption{Top 3 Hot spots in Theano}

\begin{tabular}{|c|c|c|}
\hline 
\emph{Theano Function}  & \emph{Fraction of time spent}  & \emph{Time per call to function}\tabularnewline
\hline 
\hline 
GpuAdvancedIncSubtensor1  & 81.7\%  & $4.60\e{-3}$s\tabularnewline
\hline 
GpuElemwise  & 9.2\%  & $6.93\e{-5}$s\tabularnewline
\hline 
GpuAlloc  & 1.7\%  & $1.91\e{-4}$s\tabularnewline
\hline 
\end{tabular}
\end{table}

We thus narrow down our goal to optimizing the function \emph{GpuAdvancedIncSubTensor1}.
This function typically performs an operation called \emph{advanced-indexing
}which is an operation that is typically performed while calculating
the gradient of parameter vector and updating them. In order to optimize
this, it is crucial to understand what the operation of \emph{advanced-indexing
}does. 

This operation operates takes as input 3 parameters: $W$ and $Y$
which are matrices and a vector $I$. The vector $I$ refers to
(indexes) rows in $W$. Given a row of $W$ indexed by $I$, this
operation adds the corresponding row of $Y$ to it to form the output.
This is done for each row indexed by $I$.

Having understood the operation that we are targeting to optimize,
we now outline in the next section, the optimizations we investigated
and present how the performance of this function improves with these
optimizations.

\subsection{Optimizing advanced indexing}
\label{sec:AdvancedIndexing}
To enable quick testing of advanced indexing, we wrote a standalone
script that invokes the advanced indexing operation. This enables
us to quickly test our fixes and ensure their correctness.

On examining the code for the advanced indexing, the following was
noted: 
\begin{enumerate}
\item The implementation of advanced indexing was done in Python and can
be slower than an implementation in C. Hence it would be beneficial
to rewrite the implementation in C to boost performance. 
\item The code was not highly parallelized and had a low degree of parallelism.
So we decided to not only write an implementation of advanced indexing
in C, but parallelize it as well. This was done by writing a CUDA
kernel which would perform the advanced indexing in parallel. More
specifically instead of indexing each row sequentially, each row is
indexed in parallel, and for each row, each cell in the row is added
in parallel. This greatly boosts the parallelism of the algorithm. 
\item Another micro-optimization that was tried was to also make the operation
in-place at the cost of an extra memory copy. However this was shown
to give diminishing benefits. 
\end{enumerate}
With these optimizations, the mean time taken for indexing $1000$
rows is $3.6612$ seconds($\sigma=0.14116$) compared to the baseline
where the mean time for indexing $1000$ rows was $207.59$ seconds
($\sigma=2.9652$). We also note specifically that the time per function
call to advanced indexing has reduced by a factor of $~50$.

The above speedup thus results in Advanced Indexing no longer being
the bottle-neck. In the training of the model, we do not index as
many rows(as the context size and batch sizes are smaller) and hence
the speed up in training is expected to be lower. One of the reasons
for having small batch sizes is that the model converges faster. We
in fact noted that increasing that batch size to $500$ results in
a much slower convergence rate of the model(as we will present in Section ~\ref{sec:batcheffects}).

\subsection{Speedup in training rate}
\label{sec:Speedup}
The mean rate of training is now $3742$ examples/s($\sigma=32.6496$). We thus note that we have achieved a reasonable speed up of $3-4$ times on the GPU with our optimizations and the performance on the
GPU is comparable to that on the CPU. We also note that advanced indexing is no longer a major bottleneck in the training task. On performing these optimizations, it is imperative to do a further analysis on
whether further optimizations are worth the ``bang for the buck''. This requires us to analyze exactly what is limiting the speedup. Thus in the next section we present our analysis on what is limiting the speed up on the GPU and highlight our findings below and outline the next steps.

\subsection{Analysis of limits on training performance on GPU}
\label{sec:AnalysisLimits}
We profiled the entire application(after we included all the optimizations
above) and analyzed the profile log using NVIDIA Profiler(nvprof).
From the profile, we extracted the following metrics: 
\begin{enumerate}
\item \emph{Compute Utilization: }This is the fraction of total time spent
executing on the GPU. We would like the compute utilization to be
high. If this ratio is small, this indicates that most of the time,
the GPU's are idle. For our task, we note that the Compute Utilization
is $7.4\%$ which is low 
\item \emph{Compute to Memory Op Ratio:}Out of the time spent executing
on the GPU, what fraction of time was spent doing computation to amount
of time spent in transferring data to and from device is the Compute
to Op Ratio. This ratio should be high and at least $10:1$. We noted
that this metric is $66.72$ which is high. 
\item We also note the top 2 kernels as follows:

\begin{enumerate}
\item \emph{Composite Kernel}:This kernel performs an element wise operation
on a C array which is contiguous in memory and is highly optimized
with less scope for optimization. 
\item \emph{copy\_kernel}: This is a kernel which is a part of the BLAS
Library. 
\item Also the above kernels are not expensive as they don't have expensive
operations like exponentiation etc and hence are not bottle-necks. 
\end{enumerate}
\end{enumerate}
We thus conclude that the performance of training on GPU is limited by the low compute utilization
which implies that the GPU cores are mostly idle. The low compute utilization is a fallout of the model, as we are unable to fully utilize the GPU with our current implementation. One technique of improving compute utilization in this task would be to increase the batch size of examples which implies that each batch would contain more examples and thus increase the computation on the GPU. In the next section, we present our results on this task for increasing batch sizes.

\subsection{Effect of Batch Size on training and convergence rates}
\label{sec:batcheffects}
\begin{figure}[htb!]
       \centering
        \begin{subfigure}[b]{0.8\textwidth}
                \includegraphics[width=\textwidth]{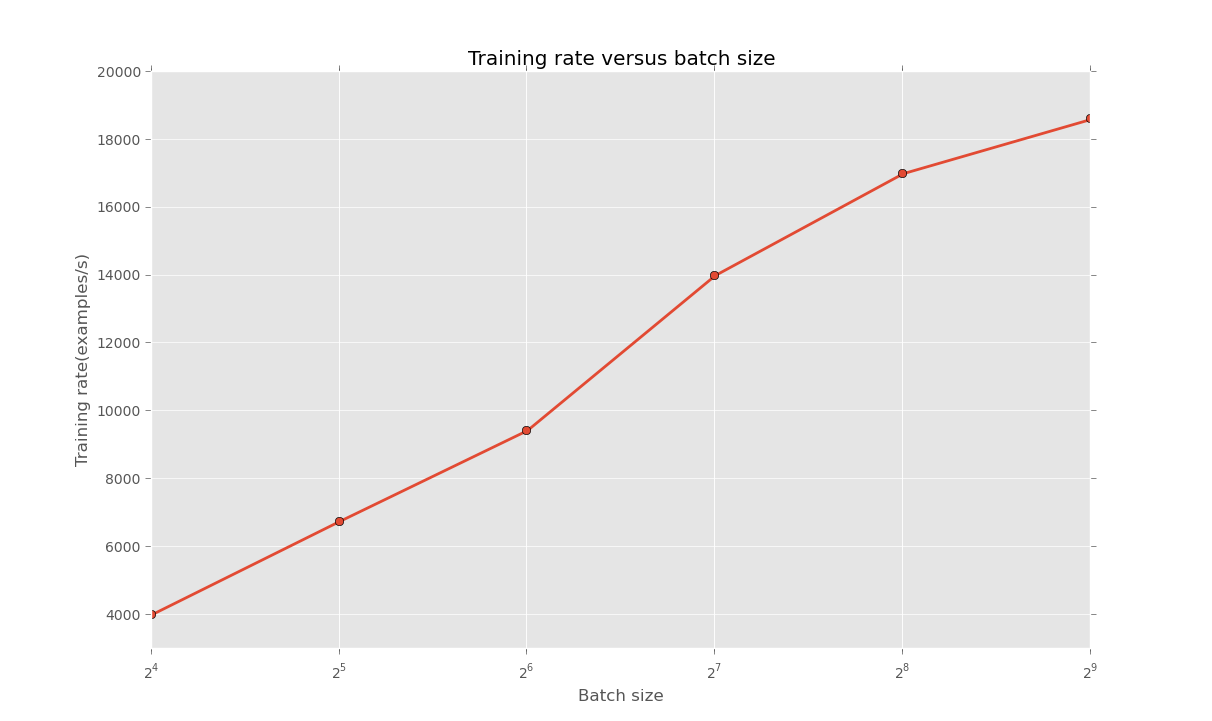}
                \caption{Effect of batch size on training rate.}
                \label{fig:batchsize_trainingrate}
        \end{subfigure}        
        \begin{subfigure}[b]{0.8\textwidth}
                \includegraphics[width=\textwidth]{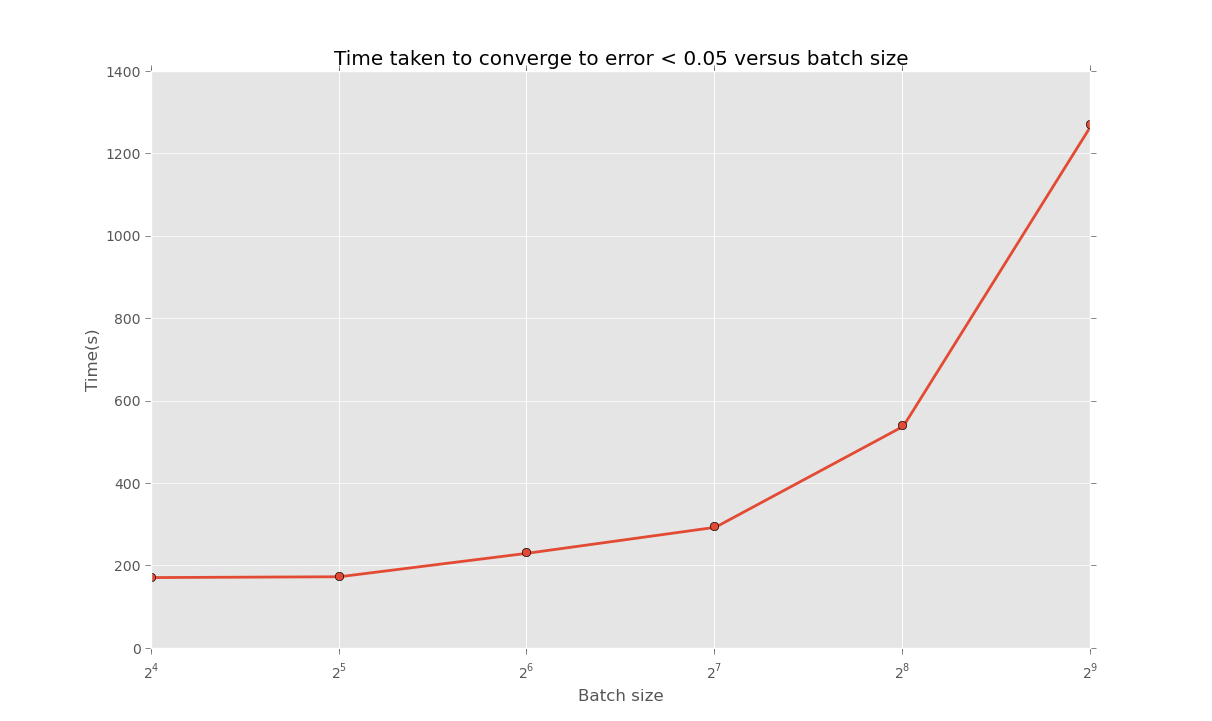}
                \caption{Effect of batch size on convergence rate}
                \label{fig:batchsize_convergence}
        \end{subfigure}        
	\caption{Effects of batch size on training and convergence rates.}
	\label{batch_fig}
\end{figure}

Currently, the batch size is set to $16$. We decided to try a range
of increasing batch sizes from $16$ to $512$ and measured the training
rate and the time taken by the model to converge to an error less
than $0.05$. We make the following observations(see Figure~\ref{batch_fig}): 
\begin{enumerate}
\item The training rate does increase as we increase the batch size. 
\item The time taken to conerge to a given error grows linearly (note the
log scale on the X-axis) as we increase the batch size 
\end{enumerate}
The observation that the convergence rate decreases as we increase the batch size can be explained by the observations made by \cite{Wilson:2003:GIB:965268.965272} that batch training is generally inefficient as compared to online training. We note that by increasing the batch size, the updates to
weights accumulate and result in larger updates. This results in the gradient descent taking unreasonably large steps thus possibly overshooting the local minima on the error surface. We thus conclude by noting
that increasing the batch size is not an effective strategy to speed up training as the model converges more slowly. In the next section, we highlight future research directions that could be investigated to speed up training.

\section{Future Work}
\label{sec:Future}
One area which could be investigated is to use the distributed algorithms for calculating gradients(using gradient descent) outlined by Jeffrey Dean et.al in \cite{NIPS2012_0598} in the model and evaluate its
performance. These algorithms update the weight vector in a distributed fashion (with updates not being synchronized). It is claimed that distributed stochastic descent performs reasonably well for the models
learnt. It would also be interesting to investigate if performance of training other models can be optimized on the GPU.It has been shown in \cite{hellinger} that Hellinger PCA can be used to learn word representations and perform competitively against word representations learnt by other models. It would be interesting to investigate whether this is amenable to good parallelization on the GPU. A second direction would be to evaluate if we could effectively use GPU's for other per-processing
tasks like annotating the Web etc.

\section{Conclusion}
\label{sec:Conclusion}
In this section, we briefly summarize our contributions below: 
\begin{enumerate}
\item We were able to significantly optimize the operation of advanced indexing
on the GPU in Theano. 
\item We showed that this boosted the performance of training on the GPU
by factor of 3-4 times and is comparable to the performance on CPU 
\item We were able to analyze exactly what the limiting factors for training
performance were on the GPU and showed that this performance was limited
by low compute utilization. 
\item We analyzed the effect of increasing batch size on the training speed
and the convergence rate and concluded that the model converges slower
for larger batch sizes even though we obtain a good speedup in training. 
\item We also contributed back our optimizations to the Open source community
so that these optimizations would benefit other members of the research
community as well. \end{enumerate}

\section{Acknowledgements}
We thank the Theano development team for helping out on code reviews for the patches submitted to Theano. 

\bibliographystyle{unsrtnat}
\renewcommand{\baselinestretch}{1}
\normalsize
\clearpage%
\phantomsection%
\addcontentsline{toc}{chapter}{\numberline{}{Bibliography}}%
\bibliography{gpu}

\begin{thebibliography}{11}
\providecommand{\natexlab}[1]{#1}
\providecommand{\url}[1]{\texttt{#1}}
\expandafter\ifx\csname urlstyle\endcsname\relax
  \providecommand{\doi}[1]{doi: #1}\else
  \providecommand{\doi}{doi: \begingroup \urlstyle{rm}\Url}\fi

\bibitem[Al-Rfou' et~al.(2013)Al-Rfou', Perozzi, and Skiena]{polyglot}
Rami Al-Rfou', Bryan Perozzi, and Steven Skiena.
\newblock Polyglot: Distributed word representations for multilingual nlp.
\newblock In \emph{Proceedings of the Seventeenth Conference on Computational
  Natural Language Learning}, pages 183--192, Sofia, Bulgaria, August 2013.
  Association for Computational Linguistics.
\newblock URL \url{http://www.aclweb.org/anthology/W13-3520}.

\bibitem[Bergstra et~al.(2010)Bergstra, Breuleux, Bastien, Lamblin, Pascanu,
  Desjardins, Turian, Warde-Farley, and Bengio]{bergstra+al:2010-scipy}
James Bergstra, Olivier Breuleux, Fr{\'{e}}d{\'{e}}ric Bastien, Pascal Lamblin,
  Razvan Pascanu, Guillaume Desjardins, Joseph Turian, David Warde-Farley, and
  Yoshua Bengio.
\newblock Theano: a {CPU} and {GPU} math expression compiler.
\newblock In \emph{Proceedings of the Python for Scientific Computing
  Conference ({SciPy})}, June 2010.
\newblock Oral Presentation.

\bibitem[Doe()]{Doe:2009:Misc}
The barracuda project.
\newblock \url{http://seqbarracuda.sourceforge.net/index.html}.

\bibitem[Fung and Mann(2005)]{Fung:2005:OPG:1101149.1101334}
James Fung and Steve Mann.
\newblock Openvidia: Parallel gpu computer vision.
\newblock In \emph{Proceedings of the 13th Annual ACM International Conference
  on Multimedia}, MULTIMEDIA '05, pages 849--852, New York, NY, USA, 2005. ACM.
\newblock ISBN 1-59593-044-2.
\newblock \doi{10.1145/1101149.1101334}.
\newblock URL \url{http://doi.acm.org/10.1145/1101149.1101334}.

\bibitem[Hinton et~al.(2006)Hinton, Osindero, and
  Teh]{Hinton:2006:FLA:1161603.1161605}
Geoffrey~E. Hinton, Simon Osindero, and Yee-Whye Teh.
\newblock A fast learning algorithm for deep belief nets.
\newblock \emph{Neural Comput.}, 18\penalty0 (7):\penalty0 1527--1554, July
  2006.
\newblock ISSN 0899-7667.
\newblock \doi{10.1162/neco.2006.18.7.1527}.
\newblock URL \url{http://dx.doi.org/10.1162/neco.2006.18.7.1527}.

\bibitem[Hinton and Salakhutdinov(2006)]{Hinton28072006}
G.~E. Hinton and R.~R. Salakhutdinov.
\newblock Reducing the dimensionality of data with neural networks.
\newblock \emph{Science}, 313\penalty0 (5786):\penalty0 504--507, 2006.
\newblock \doi{10.1126/science.1127647}.
\newblock URL \url{http://www.sciencemag.org/content/313/5786/504.abstract}.

\bibitem[Krizhevsky et~al.(2012)Krizhevsky, Sutskever, and
  Hinton]{NIPS2012_4824}
Alex Krizhevsky, Ilya Sutskever, and Geoffrey~E. Hinton.
\newblock Imagenet classification with deep convolutional neural networks.
\newblock In F.~Pereira, C.J.C. Burges, L.~Bottou, and K.Q. Weinberger,
  editors, \emph{Advances in Neural Information Processing Systems 25}, pages
  1097--1105. Curran Associates, Inc., 2012.

\bibitem[Collobert et~al.(2011)Collobert, Weston, Bottou, Karlen, Kavukcuoglu,
  and Kuksa]{collobert-2011}
Ronan Collobert, Jason Weston, L\'eon Bottou, Michael Karlen, Koray
  Kavukcuoglu, and Pavel Kuksa.
\newblock Natural language processing (almost) from scratch.
\newblock \emph{Journal of Machine Learning Research}, 12:\penalty0 2493--2537,
  Aug 2011.
\newblock URL \url{http://leon.bottou.org/papers/collobert-2011}.

\bibitem[Wilson and Martinez(2003)]{Wilson:2003:GIB:965268.965272}
D.~Randall Wilson and Tony~R. Martinez.
\newblock The general inefficiency of batch training for gradient descent
  learning.
\newblock \emph{Neural Netw.}, 16\penalty0 (10):\penalty0 1429--1451, December
  2003.
\newblock ISSN 0893-6080.
\newblock \doi{10.1016/S0893-6080(03)00138-2}.
\newblock URL \url{http://dx.doi.org/10.1016/S0893-6080(03)00138-2}.

\bibitem[Dean et~al.(2012)Dean, Corrado, Monga, Chen, Devin, Le, Mao, Ranzato,
  Senior, Tucker, Yang, and Ng]{NIPS2012_0598}
Jeffrey Dean, Greg Corrado, Rajat Monga, Kai Chen, Matthieu Devin, Quoc Le,
  Mark Mao, Marc'Aurelio Ranzato, Andrew Senior, Paul Tucker, Ke~Yang, and
  Andrew Ng.
\newblock Large scale distributed deep networks.
\newblock In P.~Bartlett, F.C.N. Pereira, C.J.C. Burges, L.~Bottou, and K.Q.
  Weinberger, editors, \emph{Advances in Neural Information Processing Systems
  25}, pages 1232--1240. 2012.
\newblock URL \url{http://books.nips.cc/papers/files/nips25/NIPS2012_0598.pdf}.

\bibitem[Lebret and Lebret(2013)]{hellinger}
R{\'e}mi Lebret and Ronan Lebret.
\newblock Word emdeddings through hellinger pca.
\newblock \emph{CoRR}, abs/1312.5542, 2013.

\end{thebibliography}
\end{document}